\pdfoutput=1
\PassOptionsToPackage{dvipsnames}{xcolor}
\documentclass[11pt]{article}

\usepackage{EMNLP2023}

\usepackage{times}
\usepackage{latexsym}

\usepackage[T1]{fontenc}

\usepackage[utf8]{inputenc}

\usepackage{microtype}

\usepackage{inconsolata}
\usepackage{algorithm}
\usepackage{algorithmic}
\usepackage{enumitem}
\usepackage{bbm}
\usepackage{csquotes}
\usepackage{multirow}
\usepackage{multicol}
\usepackage{comment}
\usepackage[dvipsnames]{xcolor}
\usepackage{graphicx}
\usepackage{caption}
\usepackage{subcaption}
\usepackage{amssymb}
\usepackage{booktabs} 
\usepackage{adjustbox}

\definecolor{crimsonglory}{rgb}{0.75, 0.0, 0.2}

\newcommand\fscore{F$_1$}
\newcommand\dcoref{\texttt{dcoref}}
\newcommand\instructgpt{\texttt{InstructGPT}}
\newcommand\chatgpt{\texttt{ChatGPT}}
\newcommand\llama{\texttt{Llama-2-Chat}}
\newcommand\codellama{\texttt{CodeLlama}}
\newcommand\longdoc{\texttt{longdoc-PC}} 
\newcommand\transferon{\texttt{TRANSFER-ON}}
 
\newcommand\spanbert{\texttt{SpanBERT}} 
\newcommand\transferen{\texttt{TRANSFER-EN}} 
\newcommand\xlmr{\texttt{XLM-R}}

\title{Are Large Language Models Robust Coreference Resolvers?}

\author{Nghia T. Le \and Alan Ritter\\
  Georgia Institute of Technology \\
    \texttt{\{nle18,alan.ritter\}@cc.gatech.edu}
\\}

\begin{document}
\maketitle
\begin{abstract}
Recent work on extending coreference resolution across domains and languages relies on annotated data in both the target domain and language \cite{xia-van-durme-2021-moving}. At the same time, pre-trained large language models (LMs) have been reported to exhibit strong zero- and few-shot learning abilities across a wide range of NLP tasks. However, prior work mostly studied this ability using artificial sentence-level datasets such as the Winograd Schema Challenge.  In this paper, we assess the feasibility of prompt-based coreference resolution by evaluating instruction-tuned language models on difficult, linguistically-complex coreference benchmarks (e.g., CoNLL-2012). We show that prompting for coreference can outperform current unsupervised coreference systems, although this approach appears to be reliant on high-quality mention detectors. Further investigations reveal that instruction-tuned LMs generalize surprisingly well across domains, languages, and time periods; yet continued fine-tuning of neural models should still be preferred if small amounts of annotated examples are available.\footnote{Our code and datasets are available at \url{https://github.com/nle18/coref-llms}}
\end{abstract}

\section{Introduction}

Entity coreference resolution aims to find all spans within an input text that refer to the same entity. As an important information extraction sub-task, coreference resolution has received considerable attention from the NLP community over the years, with recent progress driven mostly by neural coreference models \cite{lee-etal-2017-end, wu-etal-2020-corefqa, joshi-etal-2020-spanbert}. There has also been an increasing interest in the generalization of coreference systems to domains and languages beyond the popular CoNLL-2012 benchmark \cite{xia-van-durme-2021-moving, bohnet2022coreference}.  Most work on extending coreference resolution to new domains and languages relies on target language annotated data in the targeted domain, however the amount of labeled data needed to cover every possible domain in all languages is prohibitively expensive.  Meanwhile, unsupervised \cite{haghighi2010coreference} and few-shot \cite{le2022few} coreference resolution has received less attention, despite the fact that learning with less labels is desirable when adapting to new languages or domains.

Concurrently, there has been a great deal of progress on zero- and few-shot learning by prompting pre-trained language models (LMs) \cite{ouyang2022training, touvron2023llama}. There have also been attempts at evaluating pre-trained LMs coreference abilities under zero- and few-shot settings: \citet{gpt3} demonstrated that prompting GPT-3 can resolve coreference on the Winograd Schema Challenges (WSC), \citet{yang2022gpt} showed that coreference resolution was a challenging task for GPT-2 when prompted with multiple-choice templates, and \citet{agrawal2022large} successfully reframed clinical pronoun resolution as span generation.
While these studies reveal some evidence of the coreference abilities in large LMs, they either evaluate on sentence-level, artificial datasets that are designed more as an AI challenge task, use prompting methods that fail to beat reasonable baselines, or use non-standard datasets to evaluate language models' performance at coreference resolution. In contrast, the traditional dataset for coreference resolution, CoNLL-2012/OntoNotes, contains real-world document-level examples with complex linguistic annotations \cite{pradhan-etal-2012-conll}. Evaluating LMs using more realistic inputs in this setting is arguably more suitable for the evaluation of models' coreference capabilities.

\begin{figure*}[!t]
    \centering
    \includegraphics[width=1.0\textwidth]{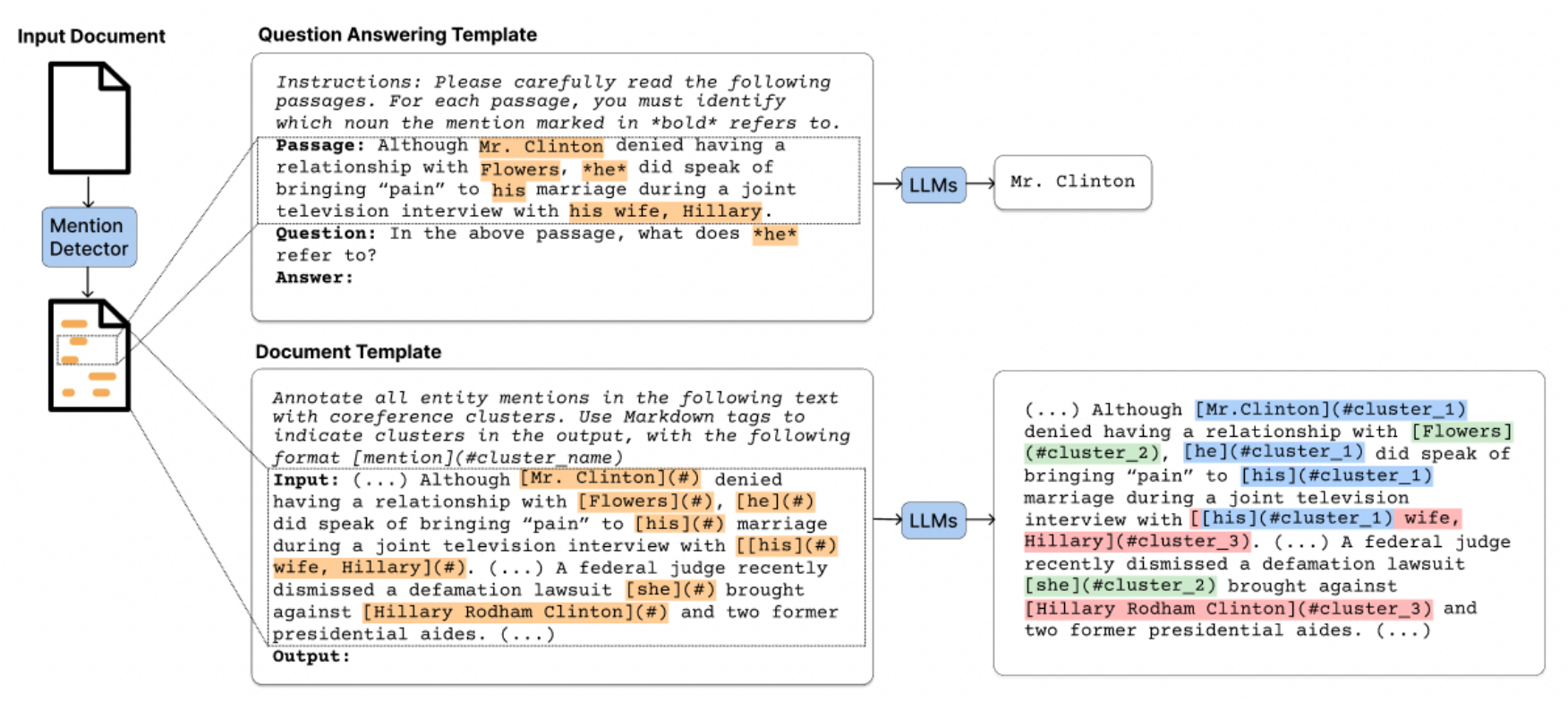}
    \caption{An example of coreference resolution with LMs prompting. Here we show two prompt templates experimented in this work: Question Answering and Document templates. In the QA template, the language model generates the answer when given a passage and an open-ended $wh$-question \cite{ouyang2022training}. In contrast, the document template marks the candidate mentions and asks the LM to annotate the cluster IDs for each mention directly within the text (represented by different colors). Both templates require a mention detector to generate candidate mentions.}
    \label{fig:approach}
\end{figure*}

In this paper, we aim to bridge the gap between coreference and language modeling literature by investigating to what extent instruction-tuned language models (e.g., InstructGPT) can perform coreference resolution via prompting. 
We show that prompting LMs is a feasible strategy for coreference resolution, outperforming previous unsupervised systems. Nonetheless, it still trails behind state-of-the-art supervised models and relies heavily on a robust mention detector. Finally, we explore the generalization ability of this approach by extending our analysis to a diverse range of domains, languages, and time periods. Our results indicate that continued learning should still be the preferred option if a large out-of-domain corpus and a few annotated in-domain documents are available. However, large instruction-tuned LMs can generalize surprisingly well across domains and languages, making them a robust option if no target language or in-domain data is available for fine-tuning. 

\section{Prompt-based Coreference Resolution}
\label{sec:prompt-coref}

Previous work in zero- and few-shot coreference resolution assumes access to candidate mentions to resolve, usually pronouns in the passage \cite{ouyang2022training, agrawal2022large}. We adopt this formulation: given a document, we assume the existence of a set of candidate mentions (gold or predicted), then prompt an autoregressive language model with handcrafted prompts, and extract the predicted coreference links (Figure \ref{fig:approach}).

Prior work applying language models to resolve co-referring entity mentions has mainly experimented with question answering prompts for pronoun resolution \cite{ouyang2022training, agrawal2022large} and demonstrated its effectiveness when comparing with other templates such as multiple-choice \cite{arora2022ama}. However, in a preliminary study  (\S \ref{appendix:prompt_format}), we found that prompting with a QA template struggled to compete with Stanford's deterministic coreference systems \cite{lee-etal-2013-deterministic}, even when providing gold mentions and few-shot guidance \cite{agrawal2022large}, or when scaling to larger LMs (achieving  61 \fscore{} when comparing to 72 \fscore{} from \citet{lee-etal-2013-deterministic}). We also experimented with an alternative document-level template that is able to elicit more coreference links than the usual QA template, achieving an 81 \fscore{}. In this template, the mentions of the input text are first marked with special tokens indicating a span to annotate (e.g., \textit{Mr. Clinton} $\rightarrow$ \textit{[Mr. Clinton](\#)}). The LM is then given instructions to annotate this marked span with the cluster ID, (e.g., \textit{[Mr. Clinton](\#)} $\rightarrow$ \textit{[Mr. Clinton](\#cluster\_1)}). Given strong results over the QA template, we used this document template for all subsequent experiments.

\section{CoNLL-2012 Experiments}
\label{sec:conll2012}

We investigate the coreference abilities of large LMs on the CoNLL-2012 benchmark \cite{pradhan-etal-2012-conll}. We found that GPT models (\instructgpt{}, \chatgpt{}, and \texttt{GPT-4}) \cite{openai2023gpt4} yield competitive results with previous unsupervised and rule-based models, while significantly outperforming them when gold mentions are provided.

\subsection{Experimental Details}

\paragraph{Dataset and Evaluation Metrics} We evaluate our approach on the traditionally benchmarked English OntoNotes 5.0 dataset \cite{ontonotes, pradhan-etal-2012-conll}, which spans seven distinct genres such as news, telephone conversations, and religious text. We follow the standard train-dev-test splits from previous work and report CoNLL \fscore{}, which averages over three coreference-based metrics MUC, B\textsuperscript{3}, and CEAF\textsubscript{$\phi_4$}.

\paragraph{Settings} We report results under two settings: predicted mentions, where only raw text is provided as input, and gold mentions, where the gold mention boundaries are provided as input. To obtain predicted mentions, we use the mentions output by \dcoref{} as input into language model prompts.

\subsection{Models}

We report performance on seven instruction-tuned language models from the LLaMa-2 \cite{touvron2023llama} and OpenAI GPT \cite{ouyang2022training} model families. We compare these models with various competitive supervised and unsupervised baselines from coreference literature.

\paragraph{Baselines} We mainly consider Stanford's deterministic resolver, which we refer to as \dcoref{} \cite{lee-etal-2013-deterministic}. This coreference resolver consists of multiple sieves, where each sieve is a set of handcrafted rules that filter out mentions. The sieves are ordered from highest to lowest precision to minimize cascading errors from previous sieves. We use the open-sourced implementation of \dcoref{} to obtain the results in this study.\footnote{https://nlp.stanford.edu/software/dcoref.html} For supervised systems, we compare to \texttt{coref-mt5}, a text-to-text approach based on mT5 from \citet{bohnet2022coreference}, which is the state-of-the-art for supervised coreference, and \texttt{SpanBERT+e2e}, a span-based neural coreference system \cite{joshi-etal-2020-spanbert}. For unsupervised baselines, we include results from \texttt{weak-SpanBERT} \cite{stolfo-etal-2022-simple}, a system that trained a SpanBERT-based coarse-to-fine architecture on \dcoref{} coreference predictions.

\paragraph{Llama 2 Models} We use models from the Llama 2 model family \cite{touvron2023llama} as the primary open-sourced language models. In particularly, we consider \llama{} 7B and 70B, as well as \codellama{} 7B and 34B. Both \llama{} and \codellama{} were instruction-tuned versions of base \texttt{Llama-2}, with \codellama{} being specifically fine-tuned on code datasets \cite{codellama}. To avoid hallucinations, we constrain the generation outputs as follows: for each given mention, we ask the model to generate the cluster ID. We then update the input sequence by appending the generated ID with the text segment between the current mention and the next mention. The process is repeated until all the mentions in the document are annotated, as in Figure \ref{fig:approach}.

\paragraph{GPT Models} We also investigate the instruction-tuned 175B GPT-3 model (\texttt{text-davinci-003}) from the InstructGPT series, which we refer to as \instructgpt{} \cite{ouyang2022training}. Previous work has reported evidence of \instructgpt{} coreference abilities via few-shot prompting \cite{agrawal2022large}. In addition, we report performance on the most recent OpenAI language models, \chatgpt{} (\texttt{gpt-35-turbo}) as well as \texttt{GPT-4} \cite{openai2023gpt4}. Due to the cost of running these models, we generate outputs using greedy decoding with a single generation per input document rather than iterative decoding as described above for the open-sourced models. 

\begin{table}[!t]
\centering
\small
\begin{tabular}{lcccc}
\toprule
System & MUC & B$^3$ & CEAF\textsubscript{4} & CoNLL \\
\midrule
& \multicolumn{3}{c}{\textit{Predicted mentions}} \\
\textcolor{darkgray}{\textit{coref-mt5}} & \textcolor{darkgray}{\textit{87.8}} & \textcolor{darkgray}{\textit{82.6}} & \textcolor{darkgray}{\textit{79.5}} & \textcolor{darkgray}{\textit{83.3}} \\
\textcolor{darkgray}{\textit{SpanBERT+e2e}} & \textcolor{darkgray}{\textit{85.3}} & \textcolor{darkgray}{\textit{78.1}} & \textcolor{darkgray}{\textit{75.3}} & \textcolor{darkgray}{\textit{79.6}} \\
\dcoref{} & 67.7 & 55.9 & 52.5 & 58.6 \\
\texttt{weak-SpanBERT} & 68.6 & 56.7 & \textbf{52.7} & 59.3 \\ 
\llama{} (70B) & 39.7 & 42.3 & 22.2 & 34.7 \\
\codellama{} (34B) & 57.5 & 40.6 & 25.3 & 41.1 \\
\instructgpt{} &  70.4 & 58.4 & 51.7 & 60.1 \\
\chatgpt{} & 66.9 & 55.5 & 46.5 & 56.3 \\
\texttt{GPT-4}  & \textbf{73.7} & \textbf{62.7} & 52.3 & \textbf{62.9} \\
\midrule 
& \multicolumn{3}{c}{\textit{Gold mentions}} \\
\dcoref{} & 81.6 & 70.0 & 67.3 & 72.9 \\
\llama{} (7B) & 19.7 & 40.2 & 22.8 & 27.6 \\
\llama{} (70B) & 58.2 & 65.7 & 34.4 & 52.8 \\
\texttt{CodeLlama} (7B)& 71.5 & 54.5 & 31.1 & 52.4 \\
\codellama{} (34B)& 75.6 & 66.5 & 43.1 & 61.7 \\
\instructgpt{} & 89.2 & 79.4 & 73.7 & 80.8 \\
\chatgpt{} & 86.2 & 79.3 & 68.3 & 77.9 \\
\texttt{GPT-4} & \textbf{93.7} & \textbf{88.8} & \textbf{82.8} & \textbf{88.4} \\
\bottomrule
\end{tabular}
\caption{Result on English OntoNotes test set for predicted mentions (top) and gold mentions (bottom). Fully supervised systems are \textcolor{darkgray}{\textit{italicized}}. Full results are reported in Table \ref{tab:main_full}.}
\label{tab:ontonotes}
\end{table}

\subsection{Results}

\begin{figure}[!t]
    \begin{subfigure}[b]{0.48\textwidth}
         \centering
        \includegraphics[width=\textwidth]{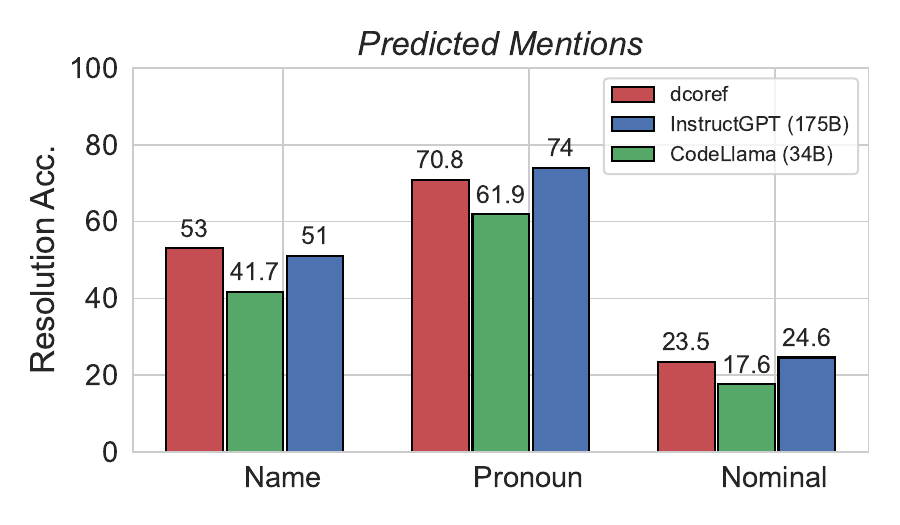}
     \end{subfigure}
     \hfill
     \begin{subfigure}[b]{0.48\textwidth}
         \centering
         \includegraphics[width=\textwidth]{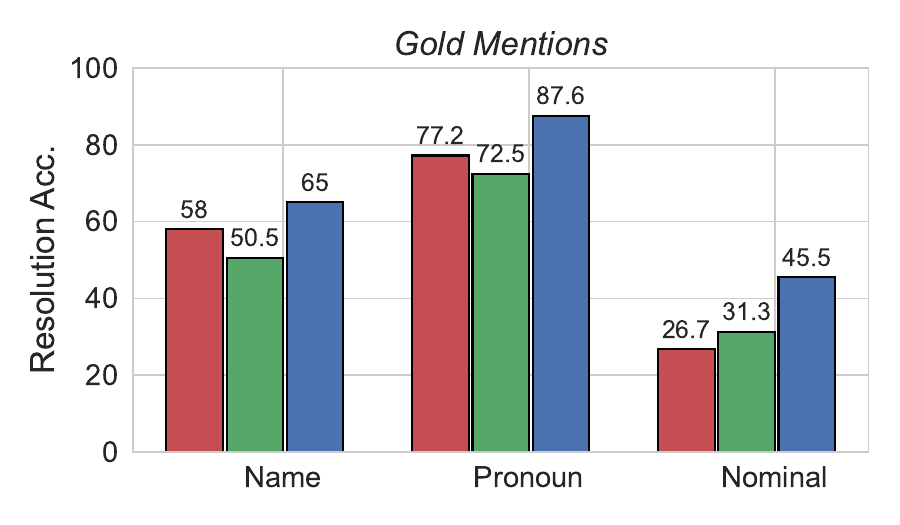}
    \end{subfigure}
    \caption{Resolution accuracy by mention types (amongst the recalled mentions) on OntoNotes dev set.}
    \label{fig:res-acc}
\end{figure}

Table \ref{tab:ontonotes} shows the results between different coreference systems. We note that prompting \instructgpt{} and \texttt{GPT-4} outperforms \texttt{weak-SpanBERT} and \dcoref{} for predicted mentions, with the performance gaps increase for gold mentions. This demonstrates the feasibility of prompting large LMs for coreference resolution, particularly in the setting where the mentions are known. However, this approach still considerably underperforms fully supervised systems. While all Llama-2 model variants underperform \dcoref{} baseline, we note that \codellama{} significantly outperforms \llama{}. \codellama{}-7B even matches the performance of \llama{}-70B.

To further understand the strengths and weaknesses of instruction-tuned LMs for coreference, we break down the results according to different \textit{resolution classes}, following \citet{lu-ng-2020-conundrums}. Specifically, for each coarse-grained mention class (named entity, pronoun, nominal), we compute the \textit{resolution accuracy}, which is the percentage of anaphors correctly linked to an antecedent (Figure \ref{fig:res-acc}). We observe that \instructgpt{} does particularly well in pronoun resolution, corroborating previous work \cite{agrawal2022large}. It struggles more for named entities and the particularly difficult nominal resolution. However, \instructgpt{} still remains competitive with \dcoref{} for these classes, with the gaps increase when gold mentions are provided. In particular, \instructgpt{} (and even \codellama{} in gold mention setting) outperforms \dcoref{} on the challenging nominal phrases case (Figure \ref{fig:res-acc}). 

\subsection{The Importance of Mention Detection}
\label{sec:mention-detection}

\begin{figure}[!t]
    \centering
\includegraphics[width=0.47\textwidth]{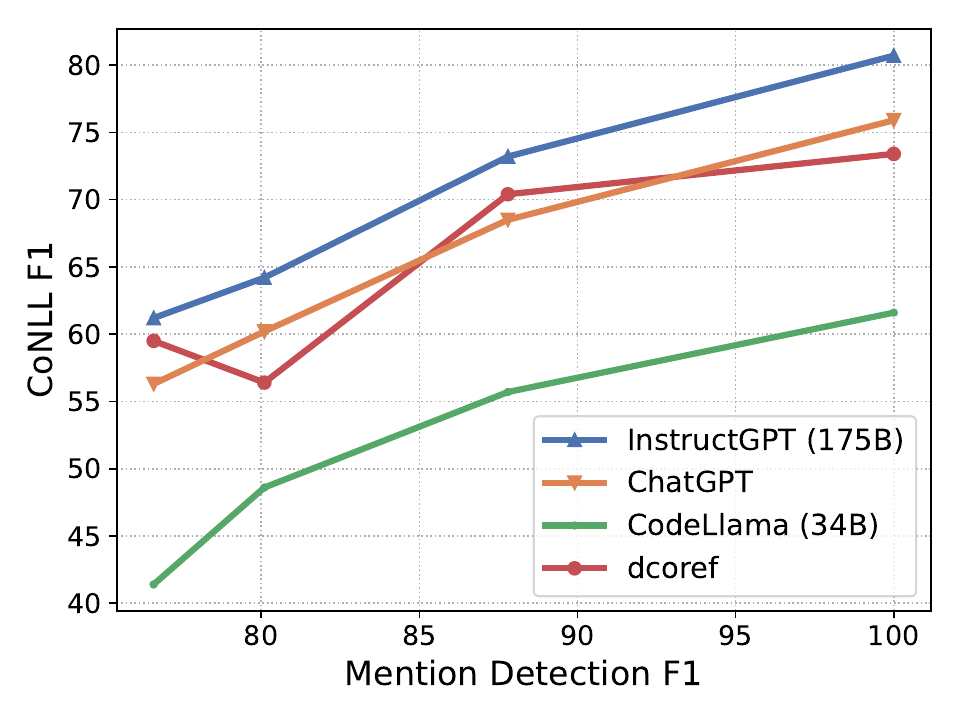}
    \caption{CoNLL \fscore{} as a function of MD \fscore{}, on OntoNotes dev set. All models were fed the same outputs from mention detection systems detailed in \S \ref{app:mention-detection}.}
    \label{fig:md-v-coref}
\end{figure}

While prompting of LMs can be competitive with previous coreference systems, the quality of candidate mentions has a considerable effect on the final performance. We quantify the importance of high-quality Mention Detection (MD) by measuring the models' performance when inputting candidate mention sets generated by different mention detectors (Figure \ref{fig:md-v-coref}).
Furthermore, we analyze the performance of \instructgpt{} when prompting for mentions with a simple template that asks it to output a list of named entities, pronouns, and nominal phrases in the input text (Table \ref{tab:mention-detection-detailed}). 

\begin{table}[!t]
\centering
\scalebox{0.92}{
\begin{tabular}{lcc}
\toprule 
Type & \textbf{\instructgpt{}} & \textbf{\dcoref{}} \\
\midrule
Name & 50.0 & \textbf{78.7} \\
Pronoun & 75.9 & \textbf{94.7} \\
Nominal & 18.7 & \textbf{52.7}\\
\midrule
MD \fscore{} & 46.5 & \textbf{76.6}\\
\bottomrule
\end{tabular}
}
\caption{Mention detection recall broken down by mention types, on OntoNotes dev set. Last row shows the overall \fscore{}. In addition to being overall worse than \dcoref{}, \instructgpt{}{} particularly struggles on recalling nominal noun phrases.}
\label{tab:mention-detection-detailed}
\end{table}

\begin{table*}[!th]
\centering
\scalebox{0.9}{
\begin{tabular}{ll}
\toprule
Mention Detection: & \textcolor{red}{[Nine years]} ago \textcolor{red}{today}, \textcolor{red}{allegations of infidelity} almost derailed \textcolor{red}{[Bill Clinton]'s} \\ 
& journey from hope to the White House. On \textcolor{red}{[January 1992]}, \textcolor{blue}{[Gennifer Flowers]} \\ 
& \textcolor{red}{claims} \textcolor{blue}{[she]} had a 12 - year affair with \textcolor{blue}{[Bill Clinton]}. \textcolor{red}{Flowers} went on \\ 
& "\textcolor{red}{[Larry King]} Live" in 1998 at the height of the \textcolor{red}{[impeachment proceedings]} \\
& against \textcolor{red}{Mr. Clinton}. \textcolor{blue}{[She]} said \textcolor{blue}{[she]} felt vindicated when \textcolor{blue}{[he]} admitted under \\
& oath that \textcolor{blue}{[he]}'d had \textcolor{red}{an affair with} \textcolor{blue}{[her]} after denying \textcolor{blue}{[it]} for years.
 \\ 
\midrule
Antecedent Linking: & Nine years ago today, \textcolor{blue}{[allegations of infidelity]\textsubscript{1}} almost derailed \textcolor{blue}{[Bill Clinton's]}\textsubscript{2} \\
(Gold Mentions) & journey from hope to the White House. On January 1992, \textcolor{blue}{[Gennifer Flowers]\textsubscript{3}} \\
& \textcolor{blue}{[claims]\textsubscript{1}} \textcolor{blue} {[she]\textsubscript{3}} had a 12 - year affair with \textcolor{blue}{[Bill Clinton]\textsubscript{2}}. \textcolor{blue}{[Flowers]\textsubscript{4}} went on \\
& “Larry King Live” in 1998 at the height of the impeachment proceedings against \\
& \textcolor{blue}{[Mr. Clinton]\textsubscript{2}}. \textcolor{blue}{[She]\textsubscript{3}} said \textcolor{blue}{[she]\textsubscript{3}} felt vindicated when \textcolor{blue}{[he]\textsubscript{2}} admitted under oath \\
& that \textcolor{blue}{[he]\textsubscript{2}}'d had \textcolor{red}{[an affair with} \textcolor{blue}{[her]\textsubscript{3}}\textcolor{red}{]\textsubscript{1}} after denying \textcolor{blue}{[it]\textsubscript{1}} for years.\\
\bottomrule
\end{tabular}
}
\caption{Qualitative examples of \instructgpt{} mention detection (top) and coreference resolution when gold mentions are given (bottom). Spans predicted by the model are wrapped around square brackets, e.g., \textcolor{red}{\textit{[Nine years]}} and \textcolor{blue}{\textit{[Bill Clinton]}}. Blue and red denote incorrect and correct predictions, respectively. \textbf{Mention Detection:} \instructgpt{} can predict most of the named entities and pronouns, but it still made numerous errors including extra entities (\textcolor{red}{\textit{Nine years}}, \textcolor{red}{\textit{January 1992}}, \textcolor{red}{\textit{Larry King}}), span errors (\textcolor{red}{\textit{Bill Clinton}} vs \textcolor{blue}{\textit{Bill Clinton's}}), and missing mentions (\textcolor{red}{\textit{Mr. Clinton}}, \textcolor{red}{\textit{Flowers}}).
\textbf{Antecedent Linking}: \instructgpt{} exhibits near perfect antecedent linking ability, with the only exception being incorrectly linking \textcolor{red}{\textit{an affair with her}} to \textcolor{blue}{\textit{allegations of infidelity}} (i.e. conflated entities error). Notably, it correctly resolved challenging cases like linking \textcolor{blue}{\textit{claims}} to \textcolor{blue}{\textit{allegations of infidelity}}. \instructgpt{} also exhibits some evidence of long-range ability when correctly resolving \textcolor{blue}{\textit{it}} to \textcolor{blue}{\textit{allegations of infidelity}}. The full example is shown in Table \ref{tab:positive-qualitative-full}, and a negative example where \instructgpt{} struggles in antecedent linking is shown in Table \ref{tab:negative-qualitative-full}.}
\label{tab:qualitative}
\end{table*}

\paragraph{\instructgpt{} consistently outperforms \dcoref{} as MD performance increases.} In general, coreference performances of all models improve as mention detection score increases. This is not surprising, as it has been similarly reported in previous work studying mention detection of neural coreference resolution systems \cite{lu-ng-2020-conundrums}. We further observe that \codellama{} underperforms while \chatgpt{} performs comparable to \dcoref{} baseline. Nonetheless, \instructgpt{} again consistently outperforms \dcoref{}, regardless of MD performance.

\paragraph{\instructgpt{} struggles with generating candidate mentions.} Table \ref{tab:mention-detection-detailed} shows that \instructgpt{} performs much worse than \dcoref{}. Further analysis by mention types shows it particularly struggles to recall nominal mentions. A qualitative example in Table \ref{tab:qualitative} demonstrates that while \instructgpt{} was able to recover a considerable portion of named entities and pronouns, it also made numerous errors, including span errors, extra entities, and missing mentions \cite{kummerfeld-klein-2013-error}. 

Given that what constitutes a mention can depend heavily on the annotation guidelines of specific datasets and domains, it may be challenging to ask a MD system to predict mentions without any labeled examples. Since Mention Detection plays a crucial role in coreference resolution \cite{wu-gardner-2021-understanding} as well as its generalizability to different domains, a high-quality mention detection appears to be a pre-requisite for prompt-based coreference resolution.  Fortunately, however, mention annotation has been shown to be much less costly than annotating full coreference chains \cite{gandhi2022mentions}.

\section{Generalization Beyond OntoNotes}

Although supervised neural models achieve superior results for coreference, they are also known to struggle when generalizing across domains, sometimes even underperforming rule-based systems \cite{moosavi-strube-2017-lexical}. As such, recent research in coreference largely focus on the generalization ability of neural models beyond the OntoNotes dataset \cite{xia-van-durme-2021-moving, gandhi2022mentions, bohnet2022coreference}. Given that large LMs are pre-trained on lots of general-purpose corpus and not optimized for a single coreference dataset, we posit that these instruction-tuned language models can also be effective at coreference domain adaptation. Therefore, we study how well instruction-tuned LMs generalize to different domains (\S \ref{ref:domains}), languages (\S \ref{ref:language}), and time periods (\S \ref{ref:temporal}). We mainly report results for \instructgpt{}, given its competitive performance on OntoNotes (\S \ref{sec:conll2012}) while being less expensive than \texttt{GPT-4}. The diverse coreference datasets considered in this analysis are given in Table \ref{tab:dataset-redux}. Since mention detection has been shown to be fairly challenging (\S \ref{sec:mention-detection}), we evaluate the experiments in this section using gold mentions.

\begin{table}[!t]
\centering
\scalebox{0.9}{
\begin{tabular}{lccc}
\toprule 
\textbf{Dataset} & \textbf{Test} & \textbf{Toks/Doc} & \textbf{\% Sing.} \\
\midrule
OntoNotes\textsuperscript{en} & 348 & 489 & 0.0 \\
\midrule
LitBank & 10 & 2105 & 19.8 \\
Character Iden. & 192 & 262 & 6.4 \\
WikiCoref & 30 & 1996 & 0.0 \\
QuizBowlCoref & 400 & 126 & 26.0 \\
\midrule
OntoNotes\textsuperscript{zh} & 218 & 412 & 0.0 \\
OntoNotes\textsuperscript{ar} & 44 & 681 & 0.0 \\
SemEval\textsuperscript{ca} & 167 & 293 & 45.9\\
SemEval\textsuperscript{nl} & 72 & 666 & 13.0\\
SemEval\textsuperscript{it} & 46 & 891 & 61.9\\
SemEval\textsuperscript{es} & 168 & 303 & 47.7\\
\midrule 
\texttt{WSJ-1989} & 56 & 632 & 0.0 \\
\texttt{WSJ-2019} & 56 & 858 & 0.0 \\
\texttt{WSJ-2023} & 56 & 688 & 0.0 \\
\bottomrule
\end{tabular}
}
\caption{Dataset statistics. The first five datasets are used as benchmarks in \citet{toshniwal-etal-2021-generalization}.We only include the number of test documents (first col.) since we evaluate the models on these datasets and did not explicitly use any train/dev data. A detailed version is shown in Table \ref{tab:dataset_details}.} 
\label{tab:dataset-redux}
\end{table}

\subsection{Can prompting LMs generalize coreference across domains?}
\label{ref:domains}

To study the robustness of our approach across domains, we use the datasets benchmarked in \citet{toshniwal-etal-2021-generalization} due to the diversity in genres (news, Wikipedia, conversations), document lengths (long vs. short), and annotation guidelines (singletons vs. non-singletons). For evaluation, we follow the annotation schema of the corresponding dataset (i.e., if the dataset contains singletons, then we also output singletons). Similar to previous work in coreference domain adaptation \cite{xia-van-durme-2021-moving, toshniwal-etal-2021-generalization}, we explore different systems where different types of source and target training data are available. Specifically, in addition to \dcoref{} as in \S \ref{sec:conll2012}, we include the \textit{trained models} \transferon{} \cite{xia-van-durme-2021-moving} and \longdoc{} \cite{toshniwal-etal-2021-generalization}, which were respectively trained on the train set of OntoNotes\textsuperscript{en} (2,802 annotated documents of newswire and religious texts) and PreCo (36,120 documents of reading comprehension examinations, collected in \citet{chen-etal-2018-preco}). \transferon{} was then further finetuned on 10 labeled documents from the target domains. Additionally, we include the \textit{pretrained encoder} \spanbert{} \cite{xia-van-durme-2021-moving} as a fine-tuning baseline (on a small amount of annotated data), where a pretrained SpanBERT encoder was not trained on a large source corpus and instead directly finetuned on 10 target documents. \footnote{Figure 1 of \citet{xia-van-durme-2021-moving}. Models summary detailed in Table \ref{tab:models}}

\paragraph{\instructgpt{} appears to be robust for coreference domain adapation.} Table \ref{tab:domains} shows the coreference domain generalization for various systems. While \instructgpt{} is competitive with \longdoc{}, it still trails behind \transferon{} considerably. This indicates that transfer learning is still a preferred method for coreference domain adaptation, particularly when a large corpus of training data and a few annotated documents in the target domain are available. On the other hand, when compared to models that were not trained on source coreference datasets such as \dcoref{} and \spanbert{}, \instructgpt{} outperforms them by a significant margin. This demonstrates the robustness of \instructgpt{} for coreference domain adaptation when using as a black-box model.

\begin{table*}[!t]
\centering
\scalebox{0.9}{
\begin{tabular}{lcccccccc}
\toprule 
\textbf{Model} & \textbf{\# Train Docs} & \textbf{ON\textsuperscript{en}} &  \textbf{LB} & \textbf{CI} & \textbf{WC} & \textbf{QBC} & \textbf{Avg.} \\
\midrule
\textit{\transferon{}} \citep{xia-van-durme-2021-moving} & 2.8k $\rightarrow$ 10 &  - & \textbf{85.0} & - & - & \textbf{85.0} & \textbf{85.0}\\
\spanbert{} \citep{xia-van-durme-2021-moving}  & 0 $\rightarrow$ 10 & - & 69.0 & - & - & 65.0 & 67.0 \\
\dcoref{} \citep{lee-etal-2013-deterministic}  & 0 $\rightarrow$ 0 & 72.9 & 55.4 & - & \underline{72.4} & 34.8 & 59.0 \\
\longdoc{} \citep{toshniwal-etal-2021-generalization} & 36k $\rightarrow$ 0 & \underline{76.8} & \underline{81.1} & \underline{66.5} & 67.0 & \underline{77.3} & 73.7 \\
\codellama{} (34B) & 0 $\rightarrow$ 0 & 61.7 & 47.8 & 58.3 & 67.9 & 58.8 & 58.9\\
\instructgpt{} & - & \textbf{80.8} & 77.0 & \textbf{72.6} & \textbf{72.9} & 68.3 & \underline{74.3}\\
\chatgpt{} & - & 77.9 & 70.8 & 67.2 & 70.8 & 69.9 & 71.3\\
\bottomrule
\end{tabular}
}
\caption{CoNLL \fscore{} on different English coreference datasets, with the macro average shown in the last column. Best result is in \textbf{bold} while the second best is \underline{underlined}. \# train docs column indicates the number of train documents from the source domain $\rightarrow$ number of train documents from target domains. \transferon{} and \longdoc{} were trained on large corpus of source examples; \transferon{} and \spanbert{} were fine-tuned on limited target examples; \dcoref{} was not trained on any corpus. Overall, \instructgpt{} exhibits strong generalization results when using out-of-the-box.}
\label{tab:domains}
\end{table*}

\begin{table}[!t]
\centering
\scalebox{0.86}{
\begin{tabular}{lccc}
\toprule 
\multirow{2}{0.05\textwidth}{\textbf{Lang.}} & \textbf{\transferen{}} & \textbf{\xlmr{}} & \multirow{2}{0.13\textwidth}{\textbf{\instructgpt{}}}\\
& \small{2.8k$\rightarrow$ 10} & \small{$0 \rightarrow$ 10} & \\
\midrule
Chinese (zh) & \underline{75.0} & 70.0 & \textbf{77.3} \\
Arabic (ar) & \textbf{80.0} & 49.0 & \underline{65.6}\\
Catalan (ca) & \textbf{52.0} & 29.0 & \underline{41.9}\\
Dutch (nl) & \textbf{71.0} & 42.0 & \underline{70.8}\\
Italian (it) & \textbf{46.0} & 25.0 & \underline{41.4}\\
Spanish (es) & \textbf{57.0} & 35.0 & \underline{42.2}\\
\bottomrule
\end{tabular}
}
\caption{CoNLL \fscore{} on the non-English portions of OntoNotes (Chinese and Arabic) and the SemEval-2010 dataset. Best result is in \textbf{bold} while the second best is \underline{underlined}. 
}
\label{tab:language}
\end{table}

\subsection{Can LMs also generalize coreference across languages?}
\label{ref:language}

To test the generalization of \instructgpt{} on resolving coreference across multiple languages, we experimented with Chinese and Arabic portions of OntoNotes and the multilingual coreference SemEval-2010 dataset \cite{recasens-etal-2010-semeval}. A notable difference between OntoNotes and SemEval-2010 is the annotations of singletons, which has led to different evaluation methods for SemEval-2010. We follow the evaluation setting of previous work for each of the evaluated languages: excluding singletons from both predicted and evaluation clusters for Chinese and Arabic, while excluding singletons from predicted set but keeping them in evaluation sets for other languages. We refer to Section 5 of \citet{bohnet2022coreference} for more discussion on this.

Similar to \S \ref{ref:domains}, we compare \instructgpt{} with neural transfer-learning models from \citet{xia-van-durme-2021-moving}, \transferen{} and \xlmr{}. Both use a pretrained \texttt{XLM-RoBERTa-large} encoder fine-tuned with 10 documents from the target language. We note that \transferen{} was previously trained on English OntoNotes before continuing training on target language, which makes it a stronger model than \xlmr{}.\footnote{Figure 6 of \citet{xia-van-durme-2021-moving}} \transferen{} and \xlmr{} correspond to \transferon{} and \spanbert{} from \S \ref{ref:domains}, respectively, with the only difference being the pretrained encoder (XLM-R vs. SpanBERT).

\paragraph{\instructgpt{} can also effectively resolve coreference across language.} From Table \ref{tab:language}, we observe similar conclusions to \S \ref{ref:domains}: continued learning using a large source corpus with a handful of annotated examples from target languages still performs the best. Nonetheless, \instructgpt{}
was able to outperform \xlmr{} across all languages, and is even on par with \transferen{} for Chinese and Dutch. This result indicates the importance of a source English coreference corpus for continued learning.

\subsection{What about different time periods?} 
\label{ref:temporal}

An interesting dimension to analyze the robustness of coreference generalization is temporal changes \cite{agarwal-nenkova-2022-temporal,liu2022conll}, since having coreference systems that can generalize beyond datasets that were created over a decade ago (e.g., OntoNotes) can be beneficial. To that end, we compare \dcoref{} and several instruction-tuned LMs on three new silver-annotated coreference datasets from different time periods: \textbf{\texttt{WSJ-1989}}, \textbf{\texttt{WSJ-2019}}, and \textbf{\texttt{WSJ-2023}}, each containing 56 Wall Street Journal articles from 1989, 2016-2019, and 2023, respectively. \texttt{WSJ-1989} is a subset of the OntoNotes dev set and thus contains gold coreference annotation. \texttt{WSJ-2019} was sampled from the RealNews dataset \cite{zellers2019grover} dated from February 2015 to February 2019, and \texttt{WSJ-2023} from the WSJ website between May and June 2023. Since these two datasets do not have coreference annotations, we used SpanBERT \cite{joshi-etal-2020-spanbert}, which was fine-tuned on the in-domain OntoNotes train set, to obtain \textit{silver annotations} for all three datasets. We then evaluate the models on these silver annotations, with mentions given as before. Further details on how we sampled and annotated these datasets are presented in \S \ref{app:temporal}.



\begin{table}[!t]
\centering
\small
\begin{tabular}{lccccc}
\toprule 
\multirow{2}{0.05\textwidth}{\textbf{Dataset}} & \texttt{1989} & \texttt{1989} & \texttt{2019} & \texttt{2023} & \multirow{2}{0.02\textwidth}{$\sigma^2$} \\
& (G) & (S) & (S) & (S) & \\
\midrule
\dcoref{} & 72.4 & 70.8 & 63.6 & 66.9 & 15.7 \\
\codellama{}\texttt{-34B} & 61.9 & 57.4 & 55.7 & 55.3 & 9.1\\
\instructgpt{} & \textbf{80.9} & \textbf{78.2} & \textbf{80.5} & \textbf{81.7} & \textbf{2.3}\\
\chatgpt{}{} & 76.8	& 75.3 & 76.7 & 74.3 & 2.5\\
\bottomrule
\end{tabular}
\caption{CoNLL \fscore{} and variance (last column) on Wall Street Journal articles from different time periods. G and S denote Gold and Silver annotation, respectively. Prompting LMs appears more robust to temporal changes than \dcoref{}.
}
\label{tab:temporal}
\end{table}

\paragraph{Prompting instruction-tuned LMs is robust to temporal changes.} Table \ref{tab:temporal} shows the results. We first observe a decrease when moving from gold to silver annotations for all models. More importantly, we see more degradation and variance in performance of \dcoref{} for the different temporal datasets, whereas the variance is less pronounced for \instructgpt{} and \chatgpt{}. While \codellama{}-34B underperforms \dcoref{} baseline, it also observes less variance when evaluated on different temporal datasets.

\section{Related Work}

\paragraph{Domain Adaptation for Coreference} Previous work has reported that neural coreference resolution trained on a single dataset struggled with out-of-domain generalization, with some performing worse than rule-based systems \cite{moosavi-strube-2017-lexical}. Several solutions to this challenge have been proposed with varying success: \citet{xia-van-durme-2021-moving} showed that continued training can help generalize to different domains and languages with as few as 10 annotated documents, and \citet{toshniwal-etal-2021-generalization} demonstrated joint training on large coreference corpora with different annotations can help neural models adapt to new domains. 

Recently, \citet{gandhi2022mentions} demonstrated that adapting mention annotations to new domains instead of the entire coreference chains is more cost-efficient while also improves domain adaptation performance. Their findings are in line with insight from analyzing different components of neural coreference systems: improving mention detection provides the largest improvement to coreference performance \cite{lu-ng-2020-conundrums, wu-gardner-2021-understanding}. We observe a similar trend with prompting \instructgpt{}.

\paragraph{Conditional Text Generation for Coreference} Research in coreference resolution has been dominated by neural span-based models that score coreference links between spans \cite{lee-etal-2017-end, joshi-etal-2020-spanbert}. Recently, a new paradigm for coreference starts to emerge: formulating coreference resolution as conditional text generation  \cite{liu+al.emnlp22, bohnet2022coreference}. Both \citet{liu+al.emnlp22} and \citet{bohnet2022coreference} fine-tuned T5-based models on sequences of structured-building actions, with the former achieving competitive results for structured prediction tasks and the latter achieving SOTA results for coreference resolution. While our work fall into this category, we are interested the intrinsic ability of the language model to resolve coreference, using an autoregressive language model on an instruction-based prompt format (as opposed to a more complex state-based format).

\paragraph{Prompting LMs for Coreference} With the success of zero-shot and few-shot prompting of large language models on various NLP benchmarks, we ask to what extent this success translates to more traditional NLP tasks like coreference resolution. \citet{Manning2020EmergentLS} shows evidence of linguistic abilities in masked LMs and \citet{blevinsPromptingLinguistics} presents a structured prompting approach that achieves strong few-shot results for sequence tagging tasks. For coreference resolution, prior work has mostly focused on few-shot learning for sentence-level, syntactically simple coreference datasets such as Winograd Schema Challenge \cite{levesque2012winograd} and for pronoun resolution on clinical data \cite{agrawal2022large}.

\section{Conclusion}

In this paper, we study how well instruction-tuned language models resolve coreference via prompting. We demonstrate the feasibility of this approach on the CoNLL-2012 benchmark, surpassing previous unsupervised systems but still underperforming state-of-the-art supervised models. Interestingly, prompting instruction-tuned LMs appears to generalize well across a wide range of domains, languages, and time periods, particularly if no training examples are given. Nonetheless, it still trails behind continued learning with a large training corpus in the source domain and a handful of annotated examples in the target domain.


\section*{Limitations}


Because OpenAI GPT models are proprietary models, we do not know whether or not OntoNotes was included in its training data.
However, at the time of writing, there is some evidence against OntoNotes data contamination. First, a previous probe that aimes to measure data contamination and memorization of OntoNotes on \chatgpt{} showed negative results. \footnote{https://hitz-zentroa.github.io/lm-contamination/blog/} Second, our experiment in \S \ref{ref:temporal} includes data sampled after the models' training cutoff date (September 2021), yet still shows a robust \fscore{}. 
Finally, the conclusions in this paper still stand regardless of whether or not these models trained on OntoNotes: (1) prompting instruction-tuned LMs is a feasible strategy for coreference resolution, and (2) although this approach has unique strengths and weaknesses, it is robust across many domains, languages, and time periods.



\bibliography{emnlp2023}
\bibliographystyle{acl_natbib}

\clearpage

\appendix

\section{Appendix}
\label{sec:appendix}

\subsection{Preliminaries on Prompt Formats}
\label{appendix:prompt_format}
\begin{figure}[!h]
    \centering
    \includegraphics[width=0.48\textwidth]{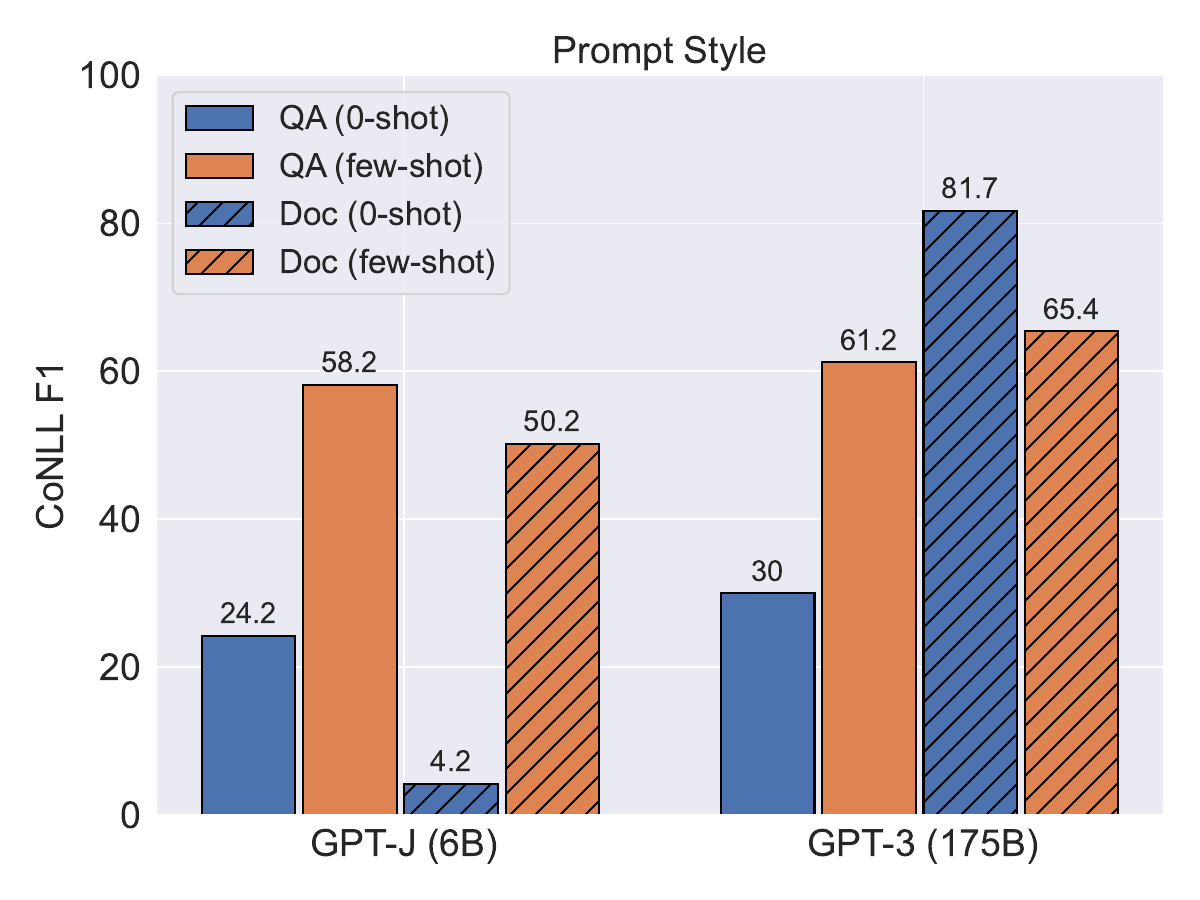}
    \caption{Results of different prompt templates for coreference on a subset of OntoNotes dev set, using gold mentions. Note that \dcoref{} achieves 71.9 \fscore{} on the same dataset.}
    \label{fig:prompt-style}
\end{figure}

\paragraph{QA Prompting for Coreference} During our preliminary studies, we experimented with different prompting approaches for coreference using QA template from previous work \cite{agrawal2022large, ouyang2022training}. However, we found that prompting \instructgpt{} this way, despite adding in-context examples to provide formatting guidance \cite{agrawal2022large}, performed consistently worse than the deterministic coreference system \dcoref{} \cite{lee-etal-2013-deterministic}. Qualitative, while this format seems effective at resolving pronouns, it would struggle with more ambiguous nominal noun phrases. For example, asking it to resolve \textit{an affair with her} in Table \ref{tab:qualitative} using QA template would yield an incorrect answer \textit{Gennifer Flowers}.

\paragraph{QA vs. Doc Template} We then experimented with the Document template (Table \ref{tab:prompts}) and found that it was more effective than the QA template at resolving coreference (Figure \ref{fig:prompt-style}). Interestingly, adding in-context examples for this template would cause \instructgpt{} to perform worse that withouto in-context examples. We further note that this Document template is loosely similar to the entity-based approach to coreference, where the model links a mention with previous clusters, as opposed to the mention-paired approach exemplified by the QA template \cite{jurafskyandmartin}. In addition, extracting the predicted clusters from the generated text is easier than other formats, as \instructgpt{} would directly annotate the text with the cluster information (We extract cluster information using a simple fuzzy string matching algorithm by comparing the output text to input text, sentence-by-sentence). 

\subsection{Mention Detection Experiments}
\label{app:mention-detection}

To experiment with different qualities of candidate mention sets, we adapting different existing methods for the task of Mention Detection: given an input document, extract all the candidate mentions from the text. For mention detection, we mainly consider the mention detector from \dcoref{} as well as the prompting of \instructgpt{} for MD using template in Table \ref{tab:prompts}. In addition, to see the effects of having high-quality mentions on \dcoref{} and \instructgpt{}, we also consider outputs from \texttt{SpanBERT-large} trained on OntoNotes train set \cite{joshi-etal-2020-spanbert} and a NER tagger with \texttt{xlm-roberta-large} \cite{conneau-etal-2020-unsupervised} trained on BIO labels adapted from OntoNotes annotations. We note that these systems are not directly comparable to each other, since they were trained on different annotatations: \texttt{SpanBERT-large} on full coreference data and \texttt{xlm-roberta-large} on non-nested MD data.

\begin{table}[!ht]
\centering
\scalebox{0.90}{
\begin{tabular}{lcccc}
\toprule 
& \textbf{Train} & \textbf{P} & \textbf{R} & \textbf{\fscore{}} \\
\midrule
\texttt{SpanBERT-large} & CR & 89.1 & 86.6 & 87.8 \\
\texttt{xlm-roberta-large} & MD & 83.3 & 76.3 & 80.1 \\
\midrule
\dcoref{} & $\varnothing$ & 75.8 & 77.4 & 76.6 \\
\instructgpt{} & - & 42.1 & 51.8 & 46.5\\
\bottomrule
\end{tabular}
}
\caption{MD results of different systems considered in Figure \ref{fig:md-v-coref}. \texttt{SpanBERT-large} was trained on full coreference (CR) data, \texttt{xlm-roberta-large} trained on mention-annotated-only (MD) OntoNotes train set, \dcoref{} was not trained on any corpus, and \instructgpt{} exact training procedures are unknown.}
\label{tab:mention-detection}
\end{table}

\subsection{Temporal Generalization for Coreference}
\label{app:temporal}

\begin{figure*}[!ht]
     \centering
     \begin{subfigure}[b]{0.48\textwidth}
         \centering
         \includegraphics[width=\textwidth]{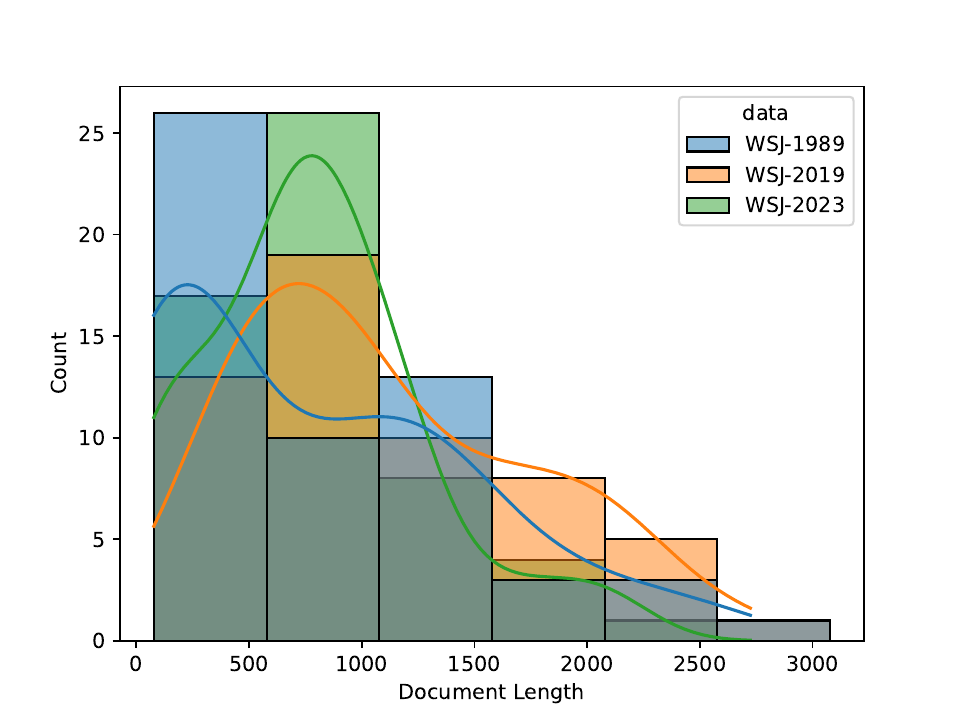}
     \end{subfigure}
     \begin{subfigure}[b]{0.48\textwidth}
         \centering
         \includegraphics[width=\textwidth]{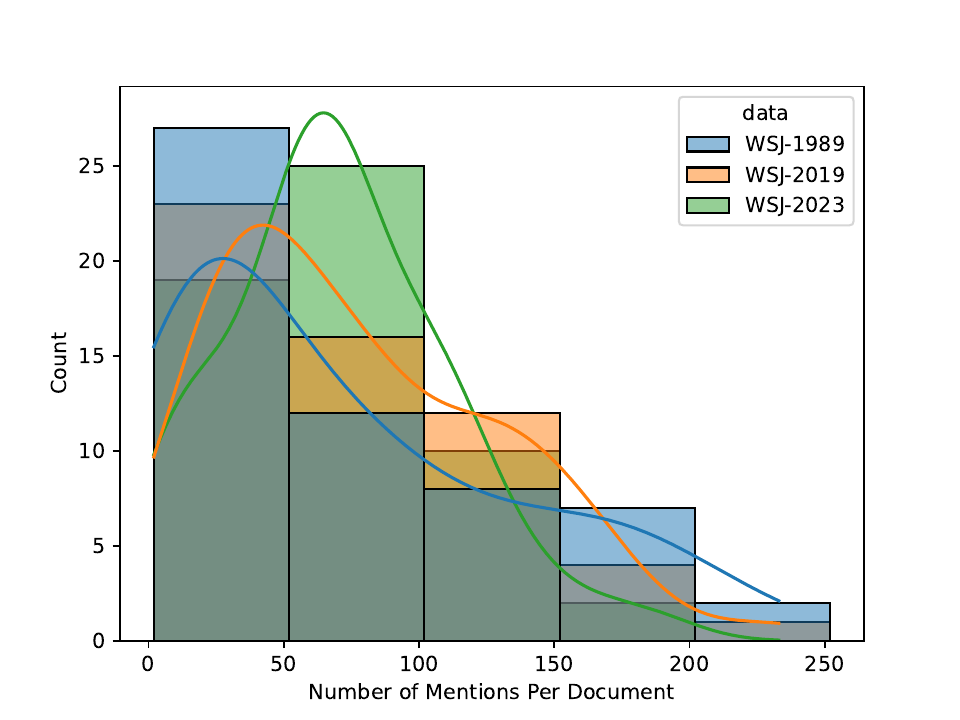}
     \end{subfigure}
     \caption{Distributions of \texttt{WSJ-1989} (blue), \texttt{WSJ-2019} (orange), and \texttt{WSJ-2023} (green) based on document length (left) and number of mentions per document (right). The number of mentions per document is measured using the silver annotations from SpanBERT \cite{joshi-etal-2020-spanbert}.
     }
    \label{fig:wsj56}
\end{figure*}

\paragraph{Data Sampling} To sample the appropriate data for this experiment, we start with the Wall Street Journal sections of the RealNews \cite{zellers2019grover} and OntoNotes dev set. We used SpanBERT \cite{joshi-etal-2020-spanbert} to label all 56 WSJ articles from OntoNotes to obtain \texttt{WSJ-1989} (CoNLL \fscore{} using SpanBERT on \texttt{WSJ-1989} is shown on Table \ref{tab:spanbert}). To create \texttt{WSJ-2019}, we first labeled all 191 WSJ articles from RealNews using SpanBERT as above. We then sampled 56 articles using stratified sampling based on two features: document length and number of mentions per document. Specifically, we partitioned the WSJ RealNews articles into bins based on document lengths (bin size = 500 tokens), and for each document-length bin we further partitioned based on the number of mentions (mention size = 50). We then sampled the appropriate number of documents (i.e., the number of \texttt{WSJ-1989} documents in each partition) for each bin to obtain \texttt{WSJ-2019}. For \texttt{WSJ-2023}, we randomly collected 56 articles from the WSJ website dated between May and June 2023 based on document lengths and topics. The distributions of three datasets are shown in Figure \ref{fig:wsj56}.

\begin{table}[!ht]
\centering
\scalebox{1.0}{
\begin{tabular}{lc}
\toprule 
Dataset & CoNLL \fscore{}\\
\midrule
OntoNotes  & 79.2 \\
\texttt{WSJ-1989} & 74.5 \\
\bottomrule
\end{tabular}
}
\caption{CoNLL \fscore{} when running SpanBERT \cite{joshi-etal-2020-spanbert} on OntoNotes dev set and \texttt{WSJ-1989}.}
\label{tab:spanbert}
\end{table}

\begin{table*}[!th]
\centering
\scalebox{0.80}{
\begin{tabular}{l}
\toprule
\textbf{Question Answering Template} \\
\midrule
\texttt{\textbf{Instructions}: Please carefully read the following passages. For each passage, you must identify} \\
\texttt{which noun the mention marked in *bold* refers to.} \\
\texttt{\textbf{Context}: In the summer of 2005, a picture that people have long been looking forward to} \\ 
\texttt{started emerging with frequency in various major Hong Kong media. With their} \\ 
\texttt{unique charm, these well-known cartoon images once again caused Hong Kong} \\
\texttt{to be a focus of worldwide attention. The world's fifth Disney park will soon} \\
\texttt{open to the public here. The most important thing about Disney is that *it* is a global brand. } \\
\texttt{\textbf{Question}: What does *it* refer to?} \\ 
\texttt{\textbf{Answer}: *it* refers to Disney.} \\
\midrule
\textbf{Document Template} \\
\midrule
\texttt{Annotate all entity mentions in the following text with coreference clusters. Use Markdown tags} \\
\texttt{to indicate clusters in the output, with the following format [mention](\#cluster\_name)} \\
\texttt{\textbf{Input}: In the summer of 2005, a picture that people have long been looking forward to started } \\
\texttt{emerging with frequency in various major [Hong Kong](\#) media . With [their](\#) unique charm,} \\
\texttt{[these well-known cartoon images](\#) once again caused [Hong Kong](\#) to be a focus of worldwide } \\
\texttt{attention. [The world's fifth [Disney](\#) park](\#) will soon open to the public here.} \\
\texttt{The most important thing about [Disney](\#) is that [it](\#) is a global brand.} \\
\texttt{\textbf{Output}: In the summer of 2005, a picture that people have long been looking forward to started}\\
\texttt{emerging with frequency in various major [Hong Kong](\#cluster\_0) media. With [their](\#cluster\_1)} \\
\texttt{unique charm, [these well-known cartoon images](\#cluster\_1) once again caused [Hong Kong](\#cluster\_0)} \\
\texttt{to be a focus of worldwide attention. [The world's fifth [Disney](\#cluster\_3) park](\#cluster\_2)}\\ 
\texttt{will soon open to the public here. The most important thing about [Disney](\#cluster\_3) is that}\\
\texttt{[it](\#cluster\_3) is a global brand.} \\
\midrule
\textbf{Mention Detection Template} \\
\midrule
\texttt{In the following text, list all named entities, pronouns, and nominal noun phrases according to}\\
\texttt{the OntoNotes conventions.} \\
\texttt{\textbf{Input:} In the summer of 2005, a picture that people have long been looking forward to started} \\
\texttt{emerging with frequency in various major Hong Kong media. With their unique charm, these}\\
\texttt{well-known cartoon images once again caused Hong Kong to be a focus of worldwide attention.}\\
\texttt{The world's fifth Disney park will soon open to the public here. The most important thing} \\
\texttt{about Disney is that it is a global brand.} \\
\texttt{\textbf{Output:}} \\
\texttt{Named Entities: Hong Kong} \\
\texttt{Pronouns: their, it, many, its, that, its, this} \\
\texttt{Nominal Noun Phrases: the summer of 2005, various major Hong Kong media, their unique charm,}\\
\texttt{the world's fifth Disney park} \\
\bottomrule
\end{tabular}
}
\caption{Examples of coreference and mention detection prompt templates used in this work.}
\label{tab:prompts}
\end{table*}


\subsection{OpenAI API Details}
\label{appendix:cost}
To maximize reproducibility, we use unconstrained greedy decoding with the temperature parameter set to 0 in all our GPT-related experiments. For \instructgpt{}, we generated approximately 18 million tokens for all our official experiments, or an equivalent of \$360. For \chatgpt{} and \texttt{GPT-4}, we generated approximately 15 million tokens (\$50) and 1 million tokens (\$60), respectively. \instructgpt{} experiments were conducted before June 2023, and \chatgpt{}/\texttt{GPT-4} experiments before December 2023.

\begin{table*}[!t]
\centering
\scalebox{0.85}{
\begin{tabular}{lcccccccccc}
\toprule
\multirow{2}{0.1\columnwidth}{System} & \multicolumn{3}{c}{MUC} & \multicolumn{3}{c}{B$^3$} & \multicolumn{3}{c}{CEAF\textsubscript{$\phi_4$}} & CoNLL \\
\cmidrule(lr){2-4}\cmidrule(lr){5-7}\cmidrule(lr){8-10}\cmidrule(lr){11-11}
& P & R & \fscore{} & P & R & \fscore{} & P & R & \fscore{} & \fscore{}\\
\midrule
& & \multicolumn{3}{c}{\textit{Predicted mentions}} & & & & & & \\
\textit{coref-mt5} \cite{bohnet2022coreference} & \textit{87.4} & \textit{88.3} & \textit{87.8} & \textit{81.8} & \textit{83.4} & \textit{82.6} & \textit{79.1} & \textit{79.9} & \textit{79.5} & \textit{83.3} \\
\textit{SpanBERT+e2e} \cite{joshi-etal-2020-spanbert} & \textit{85.8} & \textit{84.8} & \textit{85.3} & \textit{78.3} & \textit{77.9} & \textit{78.1} & \textit{76.4} & \textit{74.2} & \textit{75.3} & \textit{79.6} \\
\dcoref{} \cite{lee-etal-2013-deterministic} & 67.7 & 67.8 & 67.7 & 59.3 & 52.8 & 55.9 & 49.3 & 56.0 & 52.5 & 58.6 \\
\texttt{weak-SpanBERT} \cite{stolfo-etal-2022-simple} & 67.4 & 69.8 & 68.6 & 52.4 & 61.8 & 56.7 & 54.1 & 51.4 & \textbf{52.7} & 59.3 \\ 
\llama{} \cite{touvron2023llama} & 60.2 & 29.6 & 39.7 & 55.8 & 34.0 & 42.3 & 14.7 & 45.5 & 22.2 & 34.7 \\
\codellama{} \cite{codellama} & 54.3 & 61.0 & 57.5 & 34.3 & 49.6 & 40.6 & 22.4 & 29.1 & 25.3 & 41.1 \\
\instructgpt{} \cite{ouyang2022training} &  71.1 & 69.7 & 70.4 & 58.1 & 58.6 & 58.4 & 60.6 & 45.1 & 51.7 & 60.1 \\
\chatgpt{} \cite{openai2022chatgpt} & 67.3 & 66.5 & 66.9 & 54.3 & 56.8 & 55.5 & 43.9 & 49.5 & 46.5 & 56.3 \\
\texttt{gpt-4} \cite{openai2023gpt4}  & 73.9 & 73.5 & \textbf{73.7} & 60.8 & 64.7 & \textbf{62.7} & 49.3 & 55.7 & 52.3 & \textbf{62.9} \\
\midrule 
& & \multicolumn{3}{c}{\textit{Gold mentions}} & & & & & & \\
\dcoref{} \cite{lee-etal-2013-deterministic} & 90.0 & 74.5 & 81.6 & 84.2 & 59.7 & 70.0 & 74.4 & 61.4 & 67.3 & 72.9 \\
\texttt{llama-2-7B-chat} \cite{touvron2023llama} & 60.3 & 11.8 & 19.7 & 86.8 & 26.2 & 40.2 & 15.9 & 40.5 & 22.8 & 27.6 \\
\llama{} \cite{touvron2023llama} & 86.7 & 43.8 & 58.2 & 88.8 & 52.2 & 65.7 & 24.0 & 60.3 & 34.4 & 52.8 \\
\texttt{codellama-7B} \cite{codellama} & 72.2 & 70.7 & 71.5 & 45.2 & 68.7 & 54.5 & 30.1 & 32.1 & 31.1 & 52.4 \\
\codellama{} \cite{codellama} & 78.5 & 72.9 & 75.6 & 63.5 & 69.9 & 66.5 & 39.0 & 48.3 & 43.1 & 61.7 \\
\instructgpt{} \cite{ouyang2022training} &  89.6 & 88.9 & 89.2 & 76.0 & 89.2 & 79.4 & 84.8 & 65.2 & 73.7 & 80.8 \\
\chatgpt{} \cite{openai2022chatgpt} & 88.2 & 84.4 & 86.2 & 79.3 & 79.3 & 79.3 & 65.6 & 71.2 & 68.3 & 77.9 \\
\texttt{gpt-4} \cite{openai2023gpt4}  & 93.8 & 93.7 & \textbf{93.7} & 86.5 & 91.1 & \textbf{88.8} & 83.5 & 82.0 & \textbf{82.8} & \textbf{88.4} \\
\bottomrule
\end{tabular}
}
\caption{Result on English OntoNotes test set for predicted mentions (top) and gold mentions (bottom). Fully supervised systems are italicized.}
\label{tab:main_full}
\end{table*}

\begin{table*}[!t]
\centering
\scalebox{0.80}{
\begin{tabular}{lccccccl}
\toprule 
\textbf{Dataset} & \textbf{Lang.} & \textbf{Train} & \textbf{Dev} & \textbf{Test} & \textbf{Toks/Doc (Test)} & \textbf{\% Singletons} & \textbf{Domains} \\
\midrule
OntoNotes\textsuperscript{en} & English & 2802 & 343 & 348 & 489 & 0.0 & News, magazine, transcripts, biblical text\\
\midrule
Litbank & English & 80 & 10 & 10 & 2105 & 19.8 & Literature (Project Gutenberg) \\
Character Iden. & English & 987 & 122 & 192 & 262 & 6.4  & Movie conversations \\
WikiCoref & English & 0 & 0 &  30 & 1996 & 0.0 & Wikipedia \\
QuizBowlCoref & English & 0 & 0 &  400 & 126 & 26.0  & Trivia questions\\
\midrule
OntoNotes\textsuperscript{zh} & Chinese & 1729 & 254 &  218 & 412 & 0.0 & News, magazine \\
OntoNotes\textsuperscript{ar} & Arabic & 359 & 44 &  44 & 681 & 0.0 & News\\
SemEval\textsuperscript{ca} & Catalan & 829 & 142 &  167 & 293 & 45.9 & News \\
SemEval\textsuperscript{nl} & Dutch & 145 & 23 &  72 & 666 & 13.0 & Magazine\\
SemEval\textsuperscript{it} & Italian & 80 & 18 &  46 & 891 & 61.9 & Wikipedia, blogs, news, dialogues \\
SemEval\textsuperscript{es} & Spanish & 875 & 140 &  168 & 303 & 47.7 & News \\
\midrule
\texttt{WSJ-1989} & English & 0 & 0 & 56 & 632 & 0.0 & News (Wall Street Journal articles)\\
\texttt{WSJ-2019} & English & 0 & 0 & 56 & 858 & 0.0 & News (Wall Street Journal articles)\\
\texttt{WSJ-2023} & English & 0 & 0 & 56 & 688 & 0.0 & News (Wall Street Journal articles)\\
\bottomrule
\end{tabular}
}
\caption{Detailed statistics of datasets. Following prior work on multilingual coreference resolution \citep{bohnet2022coreference, xia-van-durme-2021-moving}, we excluded SemEval English as the data overlaps with English OntoNotes, and SemEval-2010 German due to licensing issues. We also excluded GAP, WSC, and PreCo from the benchmarks in \citet{toshniwal-etal-2021-generalization}: GAP and WSC due to the simplicity of these datasets as well as being extensively studied by previous work, and PreCo for not being able to obtain it despite contacting the authors.}
\label{tab:dataset_details}
\end{table*}

\begin{table*}[!t]
\centering
\scalebox{0.9}{
\begin{tabular}{lcc}
\toprule 
\textbf{Model} & \textbf{Prior Work} & \textbf{Description} \\
\midrule
\instructgpt{} & \citet{ouyang2022training} & pretrained on massive amount of data\\
\dcoref{} & \citet{lee-etal-2013-deterministic} & deterministic system developed on OntoNotes\textsuperscript{en}; 0-shot on target data\\
\longdoc{} & \citet{toshniwal-etal-2021-generalization} & joint training; 0-shot on target data\\
\transferon & \citet{xia-van-durme-2021-moving} & trained on OntoNotes\textsuperscript{en}; few-shot on target data\\
\spanbert & \citet{xia-van-durme-2021-moving} & pretrained on unlabeled corpus; few-shot on target data\\
\transferen & \citet{xia-van-durme-2021-moving} & trained on OntoNotes\textsuperscript{en}; few-shot on target data\\
\xlmr & \citet{xia-van-durme-2021-moving} & pretrained on unlabeled corpus; few-shot on target data\\
\bottomrule
\end{tabular}
}
\caption{Summary of models}
\label{tab:models}
\end{table*}

\begin{table*}[!th]
\centering
\scalebox{0.82}{
\begin{tabular}{ll}
\toprule
Mention Detection: & \textcolor{red}{[Nine years]} ago \textcolor{red}{today}, \textcolor{red}{allegations of infidelity} almost derailed \textcolor{red}{[Bill Clinton]'s} journey \\ 
(\instructgpt{}) & from hope to the White House. \textcolor{blue}{[Bob Glascoff]} tracks the life of \textcolor{red}{the "other woman"} \\ 
& in \textcolor{red}{[today's edition]} of "\textcolor{red}{Headliners}." On \textcolor{red}{[January 1992]}, \textcolor{blue}{[Gennifer Flowers]} \textcolor{red}{claims} \\ 
& \textcolor{blue}{[she]} had a 12 - year affair with \textcolor{blue}{[Bill Clinton]}. Although \textcolor{red}{Mr. Clinton} denied having \\ 
& a relationship with \textcolor{red}{Flowers}, \textcolor{blue}{[he]} did speak of bringing "pain" to \textcolor{blue}{[his]} marriage during \\
& a \textcolor{red}{[joint television} \textcolor{red}{interview]} with \textcolor{blue}{[his]} \textcolor{red}{wife, Hillary}. \textcolor{red}{Flowers} went on "\textcolor{red}{[Larry King]} \\ 
& Live" in 1998 at the height of the \textcolor{red}{[impeachment proceedings]} against \textcolor{red}{Mr. Clinton}. \\
& \textcolor{blue}{[She]} said \textcolor{blue}{[she]} felt vindicated when \textcolor{blue}{[he]} admitted under oath that \textcolor{blue}{[he]}'d had\\
& \textcolor{red}{an affair with} \textcolor{blue}{[her]} after denying \textcolor{blue}{[it]} for years. A \textcolor{red}{[federal judge]} recently dismissed \\
& a \textcolor{red}{[defamation lawsuit]} \textcolor{blue}{[she]} brought against \textcolor{blue}{[Hillary} \textcolor{blue}{Rodham Clinton]} and two former \\
& presidential aides. With "\textcolor{red}{Headliners}," \textcolor{red}{I}'m \textcolor{red}{[Bob Glascoff]}.
 \\ 
\midrule
Predicted Mentions: & Nine years ago \textcolor{red}{today}, \textcolor{red}{allegations of infidelity} almost derailed \textcolor{blue}{[Bill Clinton's]\textsubscript{3}} journey \\ 
(\instructgpt{}) & from hope to the White House. \textcolor{red}{Bob Glascoff} tracks the life of \textcolor{red}{the “other woman”} \\ 
& in \textcolor{red}{today's} edition of “\textcolor{blue}{[Headliners]\textsubscript{5}}.” On January 1992, \textcolor{blue}{[Gennifer Flowers]\textsubscript{6}} \textcolor{red}{claims} \textcolor{blue}{[she]\textsubscript{6}} \\ 
& had a 12-year affair with \textcolor{blue}{[Bill Clinton]\textsubscript{3}}. Although \textcolor{blue}{[Mr. Clinton]\textsubscript{3}} denied having a \\
& relationship with \textcolor{blue}{[Flowers]\textsubscript{6}}, \textcolor{blue}{[he]\textsubscript{3}} did speak of bringing “pain” to \textcolor{blue}{[his]\textsubscript{3}} marriage \\ 
& during a joint television interview with \textcolor{blue}{[his]\textsubscript{3}} \textcolor{red}{wife, Hillary}. \textcolor{blue}{[Flowers]\textsubscript{6}} went on \\ 
& \textcolor{red}{[“Larry King Live”]\textsubscript{5}} in 1998 at the height of the impeachment proceedings against \\
& \textcolor{blue}{[Mr. Clinton]\textsubscript{3}}. \textcolor{blue}{[She]\textsubscript{6}} said \textcolor{blue}{[she]\textsubscript{6}} felt vindicated when \textcolor{blue}{[he]\textsubscript{3}} admitted under oath that \\
& \textcolor{blue}{[he]\textsubscript{3}}'d had \textcolor{red}{[an affair with} \textcolor{blue}{[her]\textsubscript{6} \textcolor{red}{]\textsubscript{6}}} after denying \textcolor{blue}{[it]\textsubscript{6}} for years. A federal judge recently \\
& dismissed a defamation lawsuit \textcolor{blue}{[she]\textsubscript{6}} brought against \textcolor{red}{Hillary Rodham Clinton} \\ 
& and two former presidential aides. With “\textcolor{blue}{[Headliners]\textsubscript{5}},” \textcolor{red}{I}'m Bob Glascoff. \\
\midrule
Gold Mentions: & Nine years ago \textcolor{blue}{[today]\textsubscript{1}}, \textcolor{red}{allegations of infidelity} almost derailed \textcolor{blue}{[Bill Clinton's]}\textsubscript{3} \\
(\dcoref{}) & journey from hope to the White House. \textcolor{red}{Bob Glascoff} tracks the life of \\
 & \textcolor{red}{the “other woman”} in \textcolor{blue}{[today's]\textsubscript{1}} edition of “\textcolor{blue}{[Headliners]\textsubscript{5}}.” On January 1992, \\
& \textcolor{blue}{[Gennifer Flowers]\textsubscript{6}} \textcolor{red}{claims} \textcolor{blue} {[she]\textsubscript{6}} had a 12 - year affair with \textcolor{blue}{[Bill Clinton]\textsubscript{3}}. \\ 
& Although \textcolor{blue}{[Mr. Clinton]\textsubscript{3}} denied having a relationship with \textcolor{blue}{[Flowers]\textsubscript{6}}, \textcolor{blue}{[he]\textsubscript{3}} did\\
& speak of bringing “pain” to \textcolor{blue}{[his]\textsubscript{3}} marriage during a joint television interview with \\ 
& \textcolor{blue}{[his]\textsubscript{3}} \textcolor{red}{wife, Hillary}. \textcolor{blue}{[Flowers]\textsubscript{6}} went on 
 “Larry King Live” in 1998 at the height \\
& of the impeachment proceedings against \textcolor{blue}{[Mr. Clinton]\textsubscript{3}}. \textcolor{blue}{[She]\textsubscript{6}} said \textcolor{blue}{[she]\textsubscript{6}} felt \\ 
& vindicated when \textcolor{blue}{[he]\textsubscript{3}} admitted under oath that \textcolor{blue}{[he]\textsubscript{3}}'d had \textcolor{blue}{[an affair with} \textcolor{blue}{[her]\textsubscript{6}}\textcolor{blue}{]\textsubscript{8}} \\
& after denying \textcolor{blue}{[it]\textsubscript{8}} for years. A federal judge recently dismissed a defamation lawsuit \\
& \textcolor{blue}{[she]\textsubscript{6}} brought against \textcolor{red}{Hillary Rodham Clinton} and two former presidential\\
& aides. With “\textcolor{blue}{[Headliners]\textsubscript{5}},” \textcolor{red}{[I]\textsubscript{5}}'m Bob Glascoff.\\
\midrule
Gold Mentions: & Nine years ago \textcolor{blue}{[today]\textsubscript{1}}, \textcolor{blue}{[allegations of infidelity]\textsubscript{2}} almost derailed \textcolor{blue}{[Bill Clinton's]}\textsubscript{3} \\
(\instructgpt{}) & journey from hope to the White House. \textcolor{blue}{[Bob Glascoff]\textsubscript{4}} tracks the life of \\
 & \textcolor{blue}{[the “other woman”]\textsubscript{6}} in \textcolor{blue}{[today's]\textsubscript{1}} edition of “\textcolor{blue}{[Headliners]\textsubscript{5}}.” On January 1992, \\
& \textcolor{blue}{[Gennifer Flowers]\textsubscript{6}} \textcolor{blue}{[claims]\textsubscript{2}} \textcolor{blue} {[she]\textsubscript{6}} had a 12 - year affair with \textcolor{blue}{[Bill Clinton]\textsubscript{3}}. \\ 
& Although \textcolor{blue}{[Mr. Clinton]\textsubscript{3}} denied having a relationship with \textcolor{blue}{[Flowers]\textsubscript{6}}, \textcolor{blue}{[he]\textsubscript{3}} did\\
& speak of bringing “pain” to \textcolor{blue}{[his]\textsubscript{3}} marriage during a joint television interview with \\ 
& \textcolor{blue}{[[his]\textsubscript{3} wife, Hillary]\textsubscript{7}}. \textcolor{blue}{[Flowers]\textsubscript{6}} went on 
 “Larry King Live” in 1998 at the height \\
& of the impeachment proceedings against \textcolor{blue}{[Mr. Clinton]\textsubscript{3}}. \textcolor{blue}{[She]\textsubscript{6}} said \textcolor{blue}{[she]\textsubscript{6}} felt \\ 
& vindicated when \textcolor{blue}{[he]\textsubscript{3}} admitted under oath that \textcolor{blue}{[he]\textsubscript{3}}'d had \textcolor{red}{[an affair with} \textcolor{blue}{[her]\textsubscript{6}}\textcolor{red}{]\textsubscript{2}} \\
& after denying \textcolor{blue}{[it]\textsubscript{2}} for years. A federal judge recently dismissed a defamation lawsuit \\
& \textcolor{blue}{[she]\textsubscript{6}} brought against \textcolor{blue}{[Hillary Rodham Clinton]\textsubscript{7}} and two former presidential\\
& aides. With “\textcolor{blue}{[Headliners]\textsubscript{5}},” \textcolor{blue}{[I]\textsubscript{4}}'m Bob Glascoff.\\
\midrule
Gold Output: & Nine years ago \textcolor{blue}{[today]\textsubscript{1}}, \textcolor{blue}{[allegations of infidelity]\textsubscript{2}} almost derailed \textcolor{blue}{[Bill Clinton's]}\textsubscript{3} \\
 & journey from hope to the White House. \textcolor{blue}{[Bob Glascoff]\textsubscript{4}} tracks the life of \\
& \textcolor{blue}{[the “other woman”]\textsubscript{6}} in \textcolor{blue}{[today's]\textsubscript{1}} edition of “\textcolor{blue}{[Headliners]\textsubscript{5}}.” On January 1992, \\
& \textcolor{blue}{[Gennifer Flowers]\textsubscript{6}} \textcolor{blue}{[claims]\textsubscript{2}} \textcolor{blue} {[she]\textsubscript{6}} had a 12 - year affair with \textcolor{blue}{[Bill Clinton]\textsubscript{3}}. \\ 
& Although \textcolor{blue}{[Mr. Clinton]\textsubscript{3}} denied having a relationship with \textcolor{blue}{[Flowers]\textsubscript{6}}, \textcolor{blue}{[he]\textsubscript{3}} did\\
& speak of bringing “pain” to \textcolor{blue}{[his]\textsubscript{3}} marriage during a joint television interview with \\ 
& \textcolor{blue}{[[his]\textsubscript{3} wife, Hillary]\textsubscript{7}}. \textcolor{blue}{[Flowers]\textsubscript{6}} went on 
 “Larry King Live” in 1998 at the height \\
& of the impeachment proceedings against \textcolor{blue}{[Mr. Clinton]\textsubscript{3}}. \textcolor{blue}{[She]\textsubscript{6}} said \textcolor{blue}{[she]\textsubscript{6}} felt \\ 
& vindicated when \textcolor{blue}{[he]\textsubscript{3}} admitted under oath that \textcolor{blue}{[he]\textsubscript{3}}'d had \textcolor{blue}{[an affair with} \textcolor{blue}{[her]\textsubscript{6}}\textcolor{blue}{]\textsubscript{8}} \\
& after denying \textcolor{blue}{[it]\textsubscript{8}} for years. A federal judge recently dismissed a defamation lawsuit \\
& \textcolor{blue}{[she]\textsubscript{6}} brought against \textcolor{blue}{[Hillary Rodham Clinton]\textsubscript{7}} and two former presidential\\
& aides. With “\textcolor{blue}{[Headliners]\textsubscript{5}},” \textcolor{blue}{[I]\textsubscript{4}}'m Bob Glascoff.\\
\bottomrule
\end{tabular}
}
\caption{A qualitative examples of \instructgpt{} and \dcoref{} coreference predictions under various setting: Row 1 shows \instructgpt{} mention detection result; Row 2 shows \instructgpt{} coreference results using \dcoref{} predicted mentions; Row 3 and 4 show \dcoref{} and \instructgpt{} coreference results using gold mentions; and last row is the gold output.}
\label{tab:positive-qualitative-full}
\end{table*}

\begin{table*}[!th]
\centering
\scalebox{0.90}{
\begin{tabular}{ll}
\toprule
Mention Detection: & \textcolor{red}{[Mai Po Marshes]} adjacent to \textcolor{blue}{[Wetland Park]} is a \textcolor{red}{[major wildlife habitat]} within \textcolor{red}{[Asia]}. \\
(\instructgpt{}) & Each year, over 50,000 migratory birds fly over \textcolor{red}{[Hong Kong]'s} skyscrapers and choose to \\
& roost for winter here. As a result, \textcolor{blue}{[three different types of aviaries]} were built in \\ 
& \textcolor{blue}{[}\textcolor{red}{[Hong Kong]} \textcolor{red}{[Wetland Park]}\textcolor{blue}{]}. \textcolor{red}{These} have become the best spots to observe birds.\\ 
& Among \textcolor{red}{[common birds]}, \textcolor{red}{a rather special one} is the black-faced spoonbill. \textcolor{blue}{[It]} is \\
& \textcolor{red}{[an endangered bird species]} throughout \textcolor{red}{the [world]}. Uh-huh. Ah, there are \textcolor{red}{only about} \\
& \textcolor{red}{1,500 in the [world]}. Wow. Um, however, each year, \textcolor{red}{about [two to three hundred] of} \textcolor{blue}{[them]} \\
& come to \textcolor{blue}{[Hong Kong]} to spend the winter. Some of \textcolor{blue}{[them]}, er, have stayed in \\& \textcolor{blue}{[}\textcolor{red}{[Hong Kong]} \textcolor{red}{[Wetland Park]}\textcolor{blue}{]}. Uh-huh. So, \textcolor{red}{[our] park's} logo is unique, featuring this \\
& black-faced spoonbill , \textcolor{red}{[which]} hopefully can draw \textcolor{red}{[people's attention]}. Uh-huh.\\  
\midrule
Gold Mentions: & Mai Po Marshes adjacent to \textcolor{blue}{[Wetland Park]\textsubscript{0}} is a major wildlife habitat within Asia. \\
(\dcoref{}) & Each year, over 50,000 migratory birds fly over \textcolor{blue}{[Hong Kong's]\textsubscript{1}} skyscrapers and choose \\
& to roost for winter here. As a result, \textcolor{red}{three different types of aviaries} were built in \\
& \textcolor{blue}{[Hong Kong Wetland Park]\textsubscript{0}}. \textcolor{red}{These} have become the best spots to observe birds. Among \\
& common birds, \textcolor{blue}{[a rather special one]\textsubscript{2}} is the black-faced spoonbill. \textcolor{blue}{[It]\textsubscript{2}} is an endangered \\
& bird species throughout \textcolor{blue}{[the world]\textsubscript{3}}. Uh-huh. Ah, there are \textcolor{red}{only about 1,500 in} \textcolor{blue}{[the world]\textsubscript{3}}. \\
& Wow. Um, however, each year \textcolor{red}{about two to three hundred of} \textcolor{blue}{[them]\textsubscript{4}} come to \textcolor{blue}{[Hong Kong]\textsubscript{1}} \\
& to spend the winter. Some of \textcolor{red}{[them]\textsubscript{4}}, er, have stayed in \textcolor{blue}{[Hong Kong Wetland Park]\textsubscript{0}}. Uh-huh. \\
& So, \textcolor{blue}{[our park's]\textsubscript{0}} logo is unique, featuring this black-faced spoonbill, which hopefully can \\
& draw people's attention. Uh-huh.\\
\midrule
Gold Mentions: & Mai Po Marshes adjacent to \textcolor{red}{Wetland Park} is a major wildlife habitat within Asia. \\
(\instructgpt{}) & Each year, over 50,000 migratory birds fly over \textcolor{blue}{[Hong Kong's]\textsubscript{1}} skyscrapers and choose \\
& to roost for winter here. As a result, \textcolor{blue}{[three different types of aviaries]\textsubscript{2}} were built in \\
& \textcolor{red}{[Hong Kong Wetland Park]\textsubscript{1}}. \textcolor{blue}{[These]\textsubscript{2}} have become the best spots to observe birds. \\
& Among common birds, \textcolor{blue}{[a rather special one]\textsubscript{3}} is the black-faced spoonbill. \textcolor{blue}{[It]\textsubscript{3}} is an \\
& endangered bird species throughout \textcolor{blue}{[the world]\textsubscript{4}}. Uh-huh. Ah, there are \textcolor{red}{[only about} \\
& \textcolor{red}{1,500 in} \textcolor{blue}{[the world]\textsubscript{4}}\textcolor{red}{]\textsubscript{4}}. Wow. Um, however, each year, \textcolor{red}{[about two to three hundred of} \\
& \textcolor{red}{[them]\textsubscript{3}]\textsubscript{3}} come to \textcolor{blue}{[Hong Kong]\textsubscript{1}} to spend the winter. Some of \textcolor{blue}{[them]\textsubscript{3}}, er, have stayed in \\
& \textcolor{blue}{[Hong Kong Wetland Park]\textsubscript{1}}. Uh-huh. So, \textcolor{blue}{[our park's]\textsubscript{1}} logo is unique, featuring this black-faced \\
& spoonbill, which hopefully can draw people's attention. Uh-huh.\\
\midrule
Gold Output: & Mai Po Marshes adjacent to \textcolor{blue}{[Wetland Park]\textsubscript{2}} is a major wildlife habitat within Asia. \\
& Each year, over 50,000 migratory birds fly over \textcolor{blue}{[Hong Kong's]\textsubscript{0}} skyscrapers and choose \\
&to roost for winter here. As a result, \textcolor{blue}{[three different types of aviaries]\textsubscript{1}} were built in \\
& \textcolor{blue}{[Hong Kong Wetland Park]\textsubscript{2}}. \textcolor{blue}{[These]\textsubscript{1}} have become the best spots to observe birds. \\
& Among common birds, \textcolor{blue}{[a rather special one]\textsubscript{3}} is the black-faced spoonbill. \textcolor{blue}{[It]\textsubscript{3}} is \\
& an endangered bird species throughout \textcolor{blue}{[the world]\textsubscript{4}}. Uh-huh. Ah, there are \\
& \textcolor{blue}{[only about 1,500 in [the world]\textsubscript{4}]\textsubscript{5}}. Wow. Um, however, each year, \textcolor{blue}{[about two to three} \\
& \textcolor{blue}{hundred of [them]\textsubscript{5}]\textsubscript{6}} come to \textcolor{blue}{[Hong Kong]\textsubscript{0}} to spend the winter. Some of \textcolor{blue}{[them]\textsubscript{6}}, \\
& er, have stayed in \textcolor{blue}{[Hong Kong Wetland Park]\textsubscript{2}}. Uh-huh. So, \textcolor{blue}{[our park's]\textsubscript{2}} logo is unique, \\
& featuring this black-faced spoonbill, which hopefully can draw people's attention. Uh-huh.\\
\bottomrule
\end{tabular}
}
\caption{An example where \instructgpt{} struggles to resolve coreference, even on gold mentions. The most notable case is with nested mentions (e.g., \textcolor{red}{[about two to three hundred of [them]\textsubscript{3}]\textsubscript{3}}).}
\label{tab:negative-qualitative-full}
\end{table*}

\end{document}